\documentclass[10pt,twocolumn,letterpaper]{article}

\usepackage{cvpr}
\usepackage{times}
\usepackage{epsfig}
\usepackage{graphicx}
\usepackage{amsmath}
\usepackage{amssymb}
\usepackage{authblk}
\usepackage{subcaption}
\usepackage[breaklinks=true,bookmarks=false]{hyperref}

\cvprfinalcopy % *** Uncomment this line for the final submission

\def\cvprPaperID{****} % *** Enter the CVPR Paper ID here

\begin{document}

%%%%%%%%% TITLE
\title{Curriculum Audiovisual Learning}

\author[1]{Di Hu}
\author[2]{Zheng Wang}
\author[1]{Haoyi Xiong}
\author[2]{Dong Wang}
\author[2]{Feiping Nie}
\author[1]{Dejing Dou}
\affil[1]{Big Data Lab, Baidu Research}
\affil[2]{School of Computer Science and Center for OPTical IMagery Analysis and Learning (OPTIMAL), Northwestern Polytechnical University \authorcr {\tt\small hdui831@mail.nwpu.edu.cn}, {\tt\small zhengwangml@gmail.com}, {\tt\small xionghaoyi@baidu.com}, \authorcr {\tt\small nwpuwangdong@gmail.com}, {\tt\small feipingnie@gmail.com}, {\tt\small doudejing@baidu.com} }

\maketitle
%\thispagestyle{empty}

%%%%%%%%% ABSTRACT
\begin{abstract}
Associating sound and its producer in complex audiovisual scene is a challenging task, especially when we are lack of annotated training data. In this paper, we present a flexible audiovisual model that introduces a soft-clustering module as the audio and visual content detector, and regards the pervasive property of audiovisual concurrency as the latent supervision for inferring the correlation among detected contents. To ease the difficulty of audiovisual learning, we propose a novel curriculum learning strategy that trains the model from simple to complex scene. We show that such ordered learning procedure rewards the model the merits of easy training and fast convergence. Meanwhile, our audiovisual model can also provide effective unimodal representation and cross-modal alignment performance. We further deploy the well-trained model into practical audiovisual sound localization and separation task. We show that our localization model significantly outperforms existing methods, based on which we show comparable performance in sound separation without referring external visual supervision. Our video demo can be found at \url{https://youtu.be/kuClfGG0cFU}.
\end{abstract}

\section{Introduction}
% human audiovisual perception
Audiovisual concurrency provides potential cues for perceiving and understanding the outside world~\cite{holmes2005multisensory}. Such concurrency comes from the simple phenomena of ``Sound is produced by the oscillation of object"~\cite{hu2019deep}, and exists through our daily life, such as the talking crowd, the barking dog, the roaring machine, etc.
These inherent and pervasive correspondences provide us the reference to distinguish and correlate different audiovisual messages, then contribute to learning diversified visual appearances from their produced sounds, or perceiving various acoustic signals from their diversified sound-makers, i.e., we can visually localize the source position by hearing the sound or separate sound from chaotic acoustic scene with visual guidance.
\begin{figure}[t]
\begin{center}
   \includegraphics[width=0.9\linewidth]{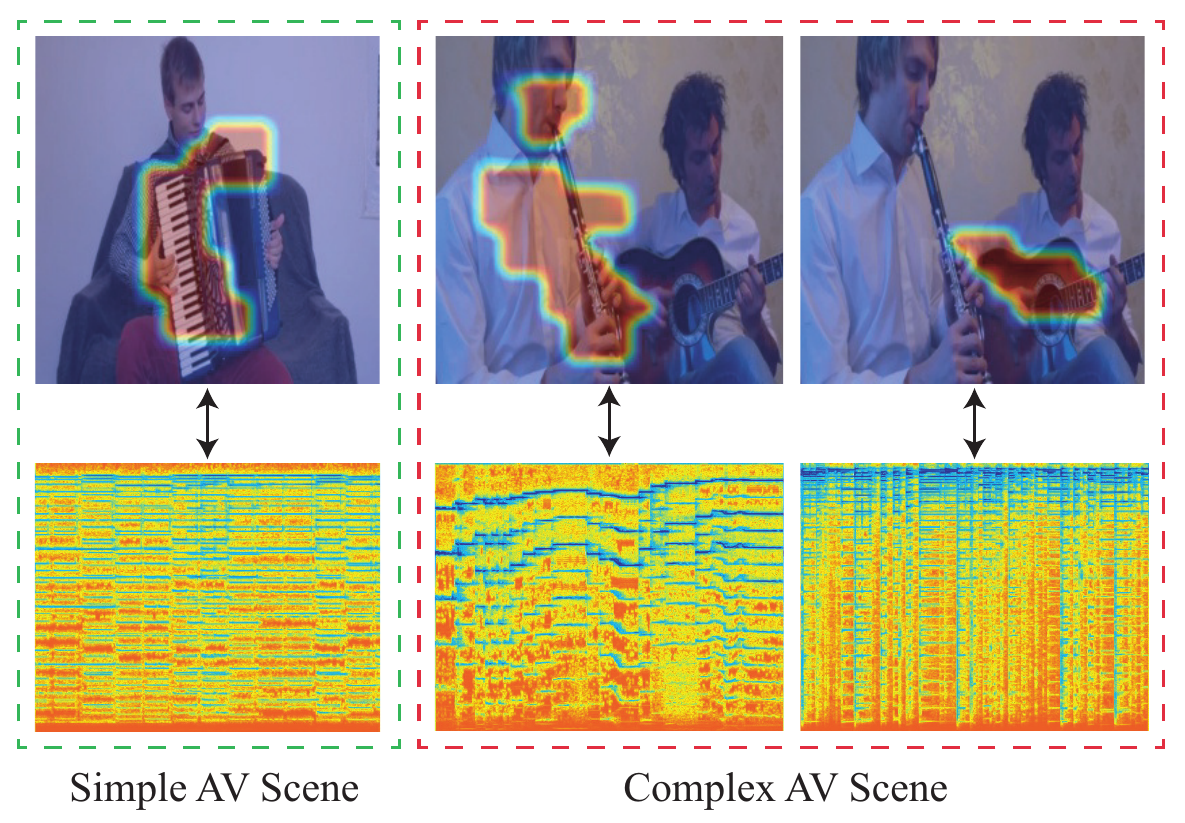}
\end{center}
   \caption{An illustration of the heterogeneous audiovisual (AV) scene complexity. The simple scene contains only one sound-source, while the complex scene contains multiple sound-sources.}
\label{illustrator}
\end{figure}

% However, unconstrained: multiple visual sound-makers,
% DMC Entities representation,  parameter issue,  alignment problems
% dis: the AV samples are handled without being distinguished; the implicit alignment across audiovisual instances
As expected, machine intelligence should also take advantage of the inherent audiovisual concurrency for possessing human-like audiovisual processing ability.  In recent years, some pioneering works have been developed to address this challenge, evolving from cross-modal knowledge transfer~\cite{aytar2016soundnet,owens2016ambient} to directly determining whether audiovisual messages are corresponding or not~\cite{arandjelovic2017look,hu2019deep,korbar2018co}. However, the learning capacity of these models is pervasively limited by the heterogeneous complexity of audiovisual scene, i.e., arbitrary number of sound-sources, as shown in Fig.~\ref{illustrator}.
% learning from scratch
On the one hand, it is easy to align sound and its visual source in the simple scene with single sound, whereas more difficult for the complex one with multiple sounds as lack of one-to-one audiovisual alignment annotations.
On the other hand, most of these existing models indiscriminately utilize these simple and complex audiovisual data, which could confuse the models when analyzing and aligning diversified audiovisual content without auxiliary annotations. Hence, we argue that differentiated audiovisual learning w.r.t. their heterogeneous complexity should be explored for achieving robust audiovisual perception.

% Application
% On other other hand, application, mostly performed as a kind of feature extractor for image/sound classification, or as pre-trained model for action classification,
% almost all the methods can localize the object, but few for object representation
% sound localization, sound separation
Further, effective audiovisual learning can reward the ability of achieving interesting cross-modal perception tasks, i.e., audiovisual sound localization and separation~\cite{owens2018audio}. Recent approaches that focus on these tasks have shown considerable performance~\cite{arandjelovic2017objects,gao2019co,senocak2018learning,zhao2018sound}.
However, these methods usually focus on simple audiovisual scene and cannot derive concrete representations for specified sound-maker (i.e., object).
Moreover, in the audiovisual sound separation task, these works normally consider that external visual knowledge from supervised training is necessary for effective separation guidance~\cite{zhao2018sound,gao2019co,zhao2019sound}, but we argue that it should be an unnecessary procedure.

% overall problems
In this paper, we strive to make a step towards to the goal of human-like audiovisual learning ability. The primary challenge is how to distinguish and align different sounds and sound-makers just with the scene-level consistency, especially when faced with the heterogeneous audiovisual complexity.
The second challenge is how to derive effective visual representations for various sound-makers without referring external knowledge, then server them to audiovisual sound localization and separation tasks.

% solution
% simple soft-kmeans clustering + alignment
% curriculum learning, complexity estimation
% deployed in two typical av-task, localization and separation. Note that the vision input to sound separation network is without supervised vision pretraining, which could still outperform most methods
% object-awareness representation
To address these two challenges, we propose to grade the heterogeneous scene into a set of audiovisual curriculum in different difficulty levels and perform differentiated audiovisual learning, from the easy ones to the hard ones. The core insight is that we can easily analyze and align the audiovisual content in the simple scenes with single sound, meanwhile it can also provide prior alignment knowledge for the learning in complex scenes.
As the audiovisual alignment is inferred by grouping and comparing the distinctive channel responses of both modalities, we can further use the aligned visual representations of sound-maker for audiovisual sound localization and separation. Concretely, our contributions are threefold:
\begin{itemize}
\item We develop a flexible audiovisual learning model that derives effective unimodal representations and infers the latent alignment between sound and sound-maker for simple and complex scene. This model imposes a soft-clustering module as the pattern detector and correlates the clustered patterns via a structured alignment objective in a space shared by both modalities.
\item We propose a novel \emph{Curriculum AudioVisual Learning} (CAVL)  strategy, where the difficulty level is determined by the number of sound-source in the scene.
Amounts of experimental results show that such simple learning strategy not only makes our model much easier to train, but also improves the learning and alignment performance of audiovisual contents. Besides, we also develop a counting model for estimating the audiovisual scene complexity.
\end{itemize}
\begin{itemize}
\item We further deploy the learned audiovisual model in cross-modal perception tasks. In audiovisual sound localization, our model shows considerable improvements over existing models. Moreover, it can provide effective visual representation for sound separation, based on which our approach has comparable performance to the methods that utilize external visual knowledge from supervised pre-training.
\end{itemize}

% 10.2
\section{Related Work}
\noindent{\textbf{Audiovisual self-supervised learning}}
% key: self-supervised learning
% dis1: wild video, complexity, inconsistency
% dis2: without alignment, inefficient
% ref: \cite{aytar2016soundnet}, \cite{harwath2016unsupervised}, \cite{owens2016ambient}
% ref: \cite{arandjelovic2017look}, \cite{hu2019deep}
As the audiovisual concurrency is inherently free of human-annotation and can provide correspondence signal of two modalities, it has drawn great attention in the self-supervised learning community~\cite{owens2018audio}, where self-supervised learning means training model from the input data itself and without human-annotation.
In the early stage, the research group in MIT firstly regards the audiovisual concurrency as the connection of cross-modal  knowledge transfer, where the student-network of one modality is supervised with the prediction of teacher-network of the other modality~\cite{aytar2016soundnet,owens2016ambient}. Recently, a novel learning criterion is proposed to directly model the audiovisual concurrency without teacher supervision~\cite{arandjelovic2017look,korbar2018co}, i.e., the audiovisual network learns to predict the sound and image from the same video or not. It is surprising that, with such simple supervision, both modality networks can learn to response to specific visual appearance and acoustic message~\cite{arandjelovic2017look,korbar2018co}. But on the other hand, as the scene-level consistency is lack of specific annotations between audiovisual components, it usually works well in the simple scenes with single sound but suffers from the defects of inefficiency and inaccuracy in the complex scene~\cite{hu2019deep}. Compared with these approaches, our method can learn to align multiple audiovisual components, even faced with complex audiovisual scene.

\noindent{\textbf{Sound localization in visual modality}}
% key: pooling operation, activation sorting, or sound separation,  from pooling to attention to universal k-means,
% dis: just a kind of verification of av correlation, do not give an efficient representation for the sound-source
% channel visualization:  \cite{zhao2018sound}, \cite{arandjelovic2017look}, \cite{owens2016ambient}
% CAM: \cite{owens2018audio}
% scalar products: \cite{arandjelovic2017objects}
% attention: \cite{senocak2018learning},
% kmeans: \cite{hu2019deep} complexity,
% they essentially perform a kind of verification of audiovisual consistency.
Visually localizing the sound-maker is a typical audiovisual perception task.  Existing approaches address this challenge mainly by correlating pixel and sound based on their concurrency, where canonical correlation analysis~\cite{Kidron2005Pixels}, embedded scalar products~\cite{arandjelovic2017objects}, attention mechanism~\cite{senocak2018learning}, and class activation map~\cite{owens2018audio} have been proposed for effectively identifying the pixels of sound. Although these methods have shown promising visualization outcome, we consider that localization results should be more than that. We can also learn effective visual representation of sound-source for further perception of vision and sound, which is expected but ignored previously.

\noindent{\textbf{Audiovisual sound separation}}
% key: mixed solo into duet for speech, or music, or motion separation
% dis: supervised/pre-trained vision
% related is \cite{owens2018audio} using pre-trained network for on-off separation
% ref: \cite{gao2018learning}, \cite{gao2019co}, \cite{zhao2019sound}, \cite{zhao2018sound}, \cite{xu2019recursive}, \cite{ephrat2018looking}
The visual information of sound-maker are considered to provide effective reference for separating the corresponding sound from complex scene~\cite{Arons1992review}.
Motivated by this, a number of approaches have been developed for achieving robust audiovisual sound separation in different types of sound, which range from speech separation~\cite{ephrat2018looking,owens2018audio}, music separation~\cite{gao2019co,zhao2019sound,xu2019recursive} to object sound separation~\cite{gao2018learning,hu2019deep}. To achieve considerable performance in realistic environment, these methods resort to the reliable visual representation of sound-maker, which are obtained from ImageNet pre-trained visual network~\cite{gao2018learning,zhao2018sound,xu2019recursive} or off-the-shelf object detector~\cite{ephrat2018looking,gao2019co}, then correlate them with the sound embeddings in a common space. Compared to these methods, our approach does not need to refer external visual knowledge trained with human annotations.

\noindent{\textbf{Curriculum learning}}
% key: for easy to hard sample
% dis: has not explored in the audiovisual scenario
% ref: \cite{bengio2009curriculum}, \cite{murali2018cassl}, \cite{korbar2018co}
Compared with a disordered arrangement, starting from easier samples or tasks then gradually increasing the difficulty level could contribute to better learning performance~\cite{Elman1993Learning,bengio2009curriculum}, which is called curriculum learning and has been widely applied in image classification~\cite{Gong2016Multi}, natural language processing~\cite{Collobert2011Natural}, speech recognition~\cite{Braun2017A}.
Different from conventional supervised learning, self-supervised learning is lack of effective human-annotation, which is more vulnerable to training order~\cite{murali2018cassl}. Recently, Korbar et al.~\cite{korbar2018co} employed curriculum learning for choosing the negative samples of audiovisual temporal synchronization.
However, there is few work focusing on how to improve the audiovisual learning performance with curriculum learning strategy, when faced with heterogeneous scene complexity.

% 10.24-25
\section{Approach}
%For an audiovisual scene with arbitrary number of sound-sources, we aims to learn disentangled audio and visual representations for each sound and sound-maker pair.To achieve this goal, we first present a model that aims to discover the contained audio and visual components, meanwhile learns to align them only with the scene-level consistency. Further, a curriculum learning strategy is developed for training this model from simple to complex audiovisual scene. Finally, we exploit the visual representation of predicted sound-maker for practical audiovisual sound localization and separation.

\subsection{Audiovisual Learning Model}
\begin{figure*}[t]
\begin{center}
   \includegraphics[width=0.9\linewidth]{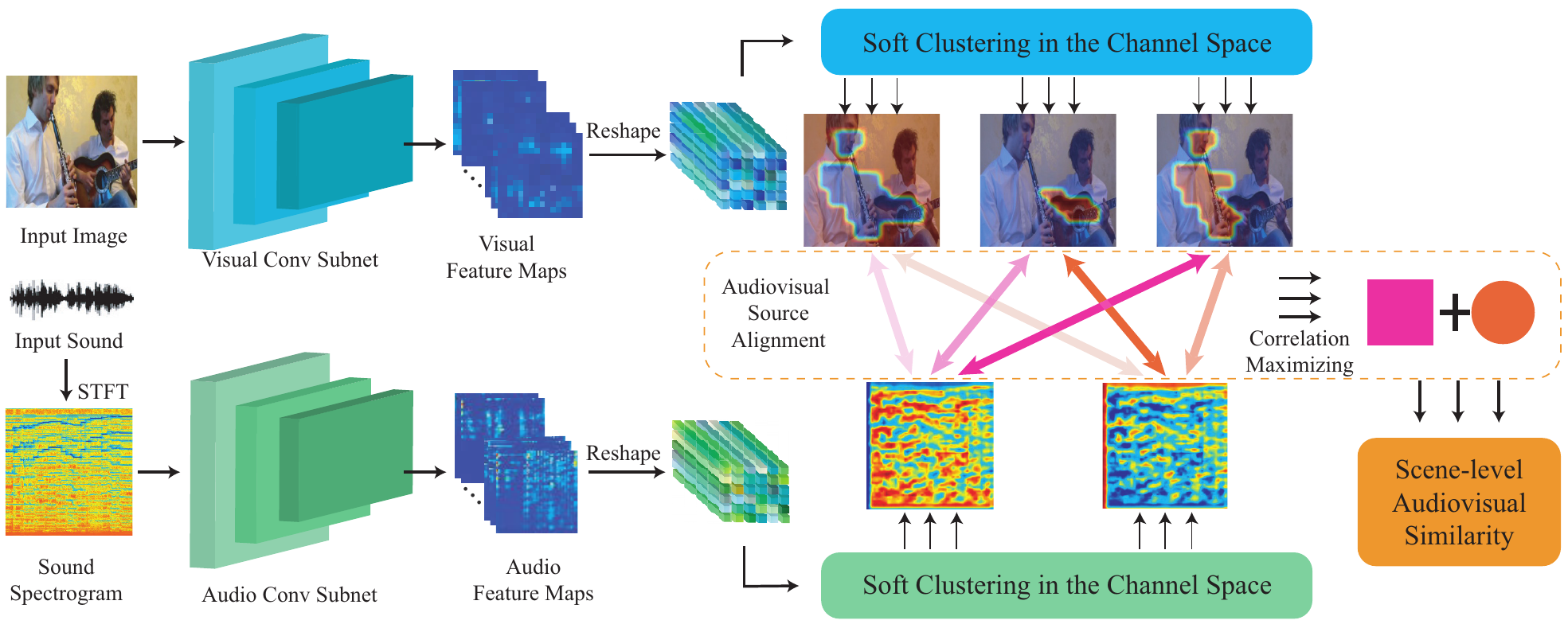}
\end{center}
   \caption{An overview of our proposed audiovisual learning model. Our model firstly represents pairwise audio and image as feature maps, then performs clustering over the reshaped maps to seek potential sound and objects, and learns to infer the alignment among these audiovisual contents. The whole model is optimized w.r.t. the scene-level similarity.}
\label{framework}
\end{figure*}

% briefly introduce the AVL model, kmeans + alignment
% one figure for cluster and alignment
Given synchronized audio and visual messages\footnote{We use sound spectrogram and image for representing audio and visual message, respectively.}, i.e., $\{ {a_1},{a_2},...,{a_n}\}$ and $\{ {v_1},{v_2},...,{v_n}\}$, which are separated from unlabelled videos  $\{ \mathcal{V}_1, \mathcal{V}_2, ..., \mathcal{V}_n \}$, we target to effectively train an audiovisual network from cold-start and make it possess the ability of generating robust unimodal representation and performing effective cross-modal perception. The whole framework is shown in Fig.~\ref{framework}.

%To address this challenge, our core insight behind is very simple, that is different objects usually take distinct appearances and produce different sounds, as the examples of clarinet and guitar in Fig.\ref{illustrator}. This naive property can help to distinguish different audio and visual components, meanwhile the concurrency between these components contributes to inferring their alignments.

\subsubsection{Learning representation via clustering}
% introduce embedded kmeans
As the filters of convolution networks have shown the property of class-relevant activation~\cite{zhou2016learning,hu2019deep}, it becomes feasible to discover and disentangle the audio and visual components by analyzing and integrating their channel representations.
Concretely, we first employ convolution networks to model each modality then embed the data into feature maps with size $H\times W \times C$, where $H$ and $W$ are the frame size and $C$ is the number of channels.
Then, these feature maps are reshaped into a set of vector $\{ {x_1},{x_2},...,{x_{H \times W}}| x_i \in R^C\}$.
Due to the distinct activation of different modal components, some of the reshaped feature vectors in the $C$-dimension channel space should take similar distribution when they describe the same modal components but dissimilar for different components.
Hence, we propose to integrate these feature vectors by performing soft K-means clustering~\cite{bauckhagelecture} in the channel space for each modality.

Formally, the objective function of soft K-means clustering~\cite{bauckhagelecture} can be formulated as
\begin{equation}\label{eq1}
J = \sum\limits_{i = 1}^{H \times W} {\sum\limits_{j = 1}^k {{w_{ij}}d\left( {{x_i},{c_j}} \right)} },
\end{equation}
where $c_j \in R^C, j=1,2,...,k$ is the $j-$cluster center, $k$ is the number of sound-sources in the audiovisual scene, $d\left( {{x_i},{c_j}} \right)$ is the Euclidean distance between current feature point $x_i$ and center $c_j$ for measuring their similarity. $w_{ij} \in [0, 1]$ is the indicator variable, which indicates the degree of assignments and can be achieved by performing softmax function over the distance of $d\left( {{x_i},{c_j}} \right)$, i.e.,
\begin{equation}\label{eq2}
{w_{ij}} = \frac{{{e^{ - \beta d\left( {{x_i},{c_j}} \right)}}}}{{\sum\limits_{l = 1}^k {{e^{ - \beta d\left( {{x_i},{c_l}} \right)}}} }},
\end{equation}
where the hype-parameter $\beta > 0$ is called stiffness parameter and controls the scalability of assignments.

Eq.~\ref{eq1} is a minimization problem about two parts, i.e., the assignments and centers. The Expectation-Maximization algorithm can be employed for solving it effectively~\cite{moon1996expectation}. In the E-step, we fix the cluster centers $C$ and update the assignment $w_{ij}$ via Eq.~\ref{eq2} . In the M-step, we fix the assignment $w_{ij}$ and re-compute the centers with the updated assignments from E-step, i.e.,
\begin{equation}\label{eq3}
{c_j} = {{\sum\limits_{i = 1}^{H \times W} {{w_{ij}}{x_i}} } \mathord{\left/
 {\vphantom {{\sum\limits_{i = 1}^{H \times W} {{w_{ij}}{x_i}} } {\sum\limits_{i = 1}^{H \times W} {{w_{ij}}} }}} \right.
 \kern-\nulldelimiterspace} {\sum\limits_{i = 1}^{H \times W} {{w_{ij}}} }}.
\end{equation}
By alternatively executing E- and M-step, we aim to find $k$ centers, each of which should correspond to certain modal component, such as specific object or sound. Meanwhile, the corresponding cluster assignment can be interpreted as a spatial-mask over feature map and indicates the location of sound-source in both modalities, as shown in Fig.~\ref{framework}.

% object means sound-maker
\subsubsection{Audiovisual alignment objective}
% introduce the audiovisual alignment
For a given audiovisual scene, although the contained sounds and objects have been described as different clustering centers, it is still difficult to directly perform alignment between them only with supervision at the entire scene level.
Fortunately, the pervasive concurrency of sound and sound-maker can help to infer the latent alignment by comparing the matching degree of different sound-object pairs, where the valid pair should possess higher matching degree.
Concretely, for each audio clustering center $c_i^a$, we aim to find proper visual clustering center (i.e., sound-maker) based on their concurrency, which can be formulated as
\begin{equation}\label{eq4}
{S_{{a}{v}}} = \sum\limits_{i = 1}^{{k_a}} {\mathop {\min }\limits_{j \in \{ 1,2,...,{k_v}\} } {{\left\| {c_i^{{a}} - c_j^{{v}}} \right\|}_2}},
\end{equation}
where $k_a$ and $k_v$ are the number of audio center and visual centers, respectively\footnote{As the visual background is usually irrelevant with sound, we set $k_v =  k_a + 1$ and use additional visual center to represent it.}. By minimizing Eq.~\ref{eq4}, each audio center is aligned to the nearest visual center, meanwhile we can also derive the similarity score of sounds and objects in the entire scene, i.e., $S_{av}$.

For an arbitrary scene, the self-supervision signal only confirms whether the audio and visual information are from the same scene (video) or not.
To effectively leverage such supervision, we employ the contrastive loss to train the audiovisual network and infer the latent alignment simultaneously, which has shown the property of consistency and robust in two-stream network optimization~\cite{koch2015siamese,korbar2018co}.
Concretely, the contrastive loss is written as
\begin{equation}\label{eq5}
\begin{aligned}
    L_{av} = \frac{1}{{2n}}&\sum\limits_{i,j = 1}^n ({\delta _{i = j}}S_{{a_i}{v_j}}^2 + \\
    &(1 - {\delta _{i = j}})\max {{(margin - {S_{{a_i}{v_j}}},0)}^2})
\end{aligned}
\end{equation}
where $a_i$ and $v_j$ stand for the sound and image from scene $i$ and $j$, respectively. $\delta _{i = j}$ is an indicator of each sound-image pair, i.e., $\delta _{i = j} = 1$ if $i=j$, otherwise $0$.
In practice, the negative samples of $i\ne j$ are randomly sampled from the training set. Generally, Eq.~\ref{eq5} encourages the audiovisual network to have higher matching confidence for the aligned sound-image pair than the mismatched ones by introducing the hyper-parameter of $margin$.

\subsection{Curriculum Learning}
\subsubsection{Curriculum Procedure}
% briefly introduce the curriculum learning motivation and settings
Usually, the audiovisual scenes in the wild contain different amounts of sound-sources, we find that directly performing audiovisual learning with these data will make the model very difficult to optimize and also lower the alignment performance.
To settle this problem, we propose to train the audiovisual model step by step, which is about starting from simple scene then gradually increasing the difficulty level, where the number of sound-sources is considered as the reference for audiovisual scene complexity.
Intuitively, for the simple scene with single sound-source, it is easy to visually localize the sound-maker from background then align it to the unique sound, such as the example of accordion in Fig.~\ref{illustrator}.
By contrast, we find that if training with complex audiovisual scenes (e.g., with three sound-sources) from the beginning, the model will get much lower convergence speed and worse results.
While, model trained with simple scenes can contribute prior knowledge for distinguishing different sound-makers and sounds, and also provides the reference for alignment . Therefore, the audiovisual learning model can be further optimized with the complex scene, which leads to better results.

In practice, to effectively perform curriculum learning, all the audiovisual data have been sorted from simple to complex before training, according to the number of sound-sources in the scene. For different learning stages, the cluster number is accordingly set to the number of sources, e.g., $k_a=1$ and $k_v=2$ for the scene with single source. Based on these graded audiovisual data, we can train the audiovisual learning model in a curriculum fashion.

\subsubsection{Complexity Estimation}
% estimating the number of sound-source
Since the audiovisual scene complexity is crucial for curriculum training, it is worth to learn to model and estimate the number of sound-source in a given scene.
Formally, the discrete probability distribution of Poisson for counting data $y_i$ is given by
%\begin{equation}\label{eq6}
$P({Y_i} = {y_i}) = \frac{{{e^{ - {\lambda _i}}}\lambda _i^{{y_i}}}}{{{y_i}!}},{y_i} = 1,2,...,$
%\end{equation}
where $\lambda _i$ is interpreted as the expected number of events in the interval and ${y_i}!$  is the factorial of $y_i$.
In this task, $y_i$ is performed as the number of sound source in the audiovisual scene.
Consequently, we propose to model $\lambda _i$ as a function of the input sound $a_i$ by the audio network, which is written as ${\lambda _i} = f(a_i)$. The function $f(\cdot)$ means the counting network of sound-sources.
By taking the negative log-likelihood w.r.t. the Poisson distribution, we can have the Poisson regression loss
\begin{equation}\label{eq8}
{L_p} = \sum\limits_{i = 1}^n {(f(a_i) - {y_i}\log f(a_i))+ {\rm{ln}}({y_i}!)},
\end{equation}
where the term of ${\rm{ln}}({y_i}!)$ can be ignored, as it is a constant to the model training.
After training the counting network, we can estimate the scene complexity by identifying the number that holds the maximal probability.

\subsection{Audiovisual Perception}
%Effective audiovisual learning can help to perform cross-modal inference, such as localizing the visual position of sound-source or separating the object sound from complex environment.
\subsubsection{Localizing sounds in visual modality}
% sound localization
Considering that the audiovisual learning model has learned to align objects and sounds in the training phase, we can directly identify the potential object which produces given sound by comparing their similarity, i.e.,
\begin{equation}\label{eq8}
c_{source}^v = \mathop {\arg \min }\limits_{j = 1,2,...,{k_v}} {\left\| {c_{source}^a - c_j^v} \right\|_2}.
\end{equation}
For the clustering center of sound-sources $c_{source}^a$ , we compare it with all the visual centers $\{ {c_1^v},{c_1^v},...,{c_{k_v}^v}\}$ and select the closest one as visual representation of sound source.
As the corresponding assignment $w_{\cdot source}$   indicates the correlation between all the visual feature vectors and  $c_{source}^v$  ,we can reshape it back to the size of $H \times W$ and regard it as the location mask of sound-maker to achieve visual localization. To better visualize the object position, we can also further resize the assignment to the size of input image.

\subsubsection{Audiovisual sound separation}
% sound separation
% one figure with mask/center
To validate the effectiveness of inferred object representation further, we propose to perform sound separation based on visual guidance.
The representative audiovisual separation network in~ \cite{ephrat2018looking,gao2019co} is adopted, as shown in Fig.~\ref{separation}.
Concretely, the separation network takes the visual clustering center (i.e., $c_{i - source}^v$) as the sound-maker representation in $i-$scene, and targets to separate its produced sound from the mixed audio signal.
Alternatively, we can also use the assignment $w_{\cdot source}$ to point out the location of sound-maker, and regard it as the object mask over the visual feature maps. Then, a sound-maker-awareness max-pooling can be performed over the masked feature maps to obtain robust object representation.

\begin{figure}[t]
\begin{center}
   \includegraphics[width=0.9\linewidth]{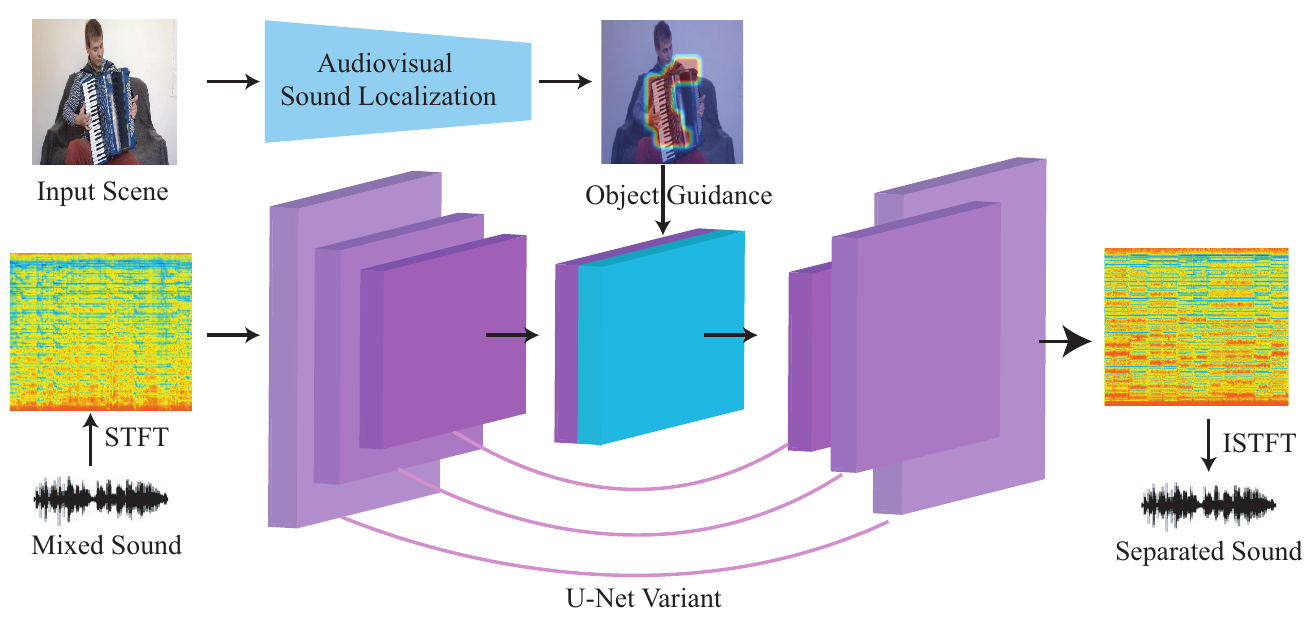}
\end{center}
\vspace{-0.2cm}
   \caption{An illustration of the audiovisual sound separation model. This model uses automatically localized object as the visual guidance for separating its produced sound from the mixed audio signal.}
\label{separation}
\end{figure}

% we use the conventional sound separation network of U-NET variant.
A variant of U-Net~\cite{ronneberger2015u} is used to perform sound-source separation, similar to~\cite{gao2019co,zhao2018sound}. The network takes the spectrogram of $i-$mixed sound $a_i^{mix}$ as the input, then encodes it into audio feature maps via stacked convolution layers. The replication and tiling operation is performed over the visual representation to match the size of embedded audio feature maps. Then we concatenate these two modalities, and feed them into stacked up-convolution layer to generate a spectrogram mask. The separation loss is written as
\begin{equation}\label{eq9}
{L_s} = \sum\limits_{i = 1}^n {{{\left\| {{M_i} - g\left( {{{a}_i^{mix}},c_{i - source}^v} \right)} \right\|}_1}} ,
\end{equation}
where $g(\cdot)$ means the audiovisual sound separation network and $M_i$ is the mask of the spectrogram magnitudes of the target sound $a_i$ and mixed sound ${{a}_i^{mix}}$ , i.e.,
${M_i} = {{{m^{{a_i}}}} \mathord{\left/
 {\vphantom {{{m^{{a_i}}}} {{m^{a_i^{mix}}}}}} \right.
 \kern-\nulldelimiterspace} {{m^{a_i^{mix}}}}}$. With the masked spectrogram, we can use \emph{Inversed Short-Time Fourier Transform} (ISTFT) to produce separated sound signal w.r.t. specific sound-maker.

\section{Experiments}
\subsection{Datasets}
\noindent{\textbf{AudioSet-Balanced}}
Audioset\cite{gemmeke2017audio} is an audio event dataset, which consists of 2,084,320 human-annotated 10-second video clips. These clips are collected from YouTube, therefore many of which are in poor-quality and contain multiple sound-sources. A hierarchical ontology of 632 event classes is employed to annotate these data, which means that the same sound could be annotated as different labels. For example, the sound of barking is annotated as \emph{Animal}, \emph{Pets}, and \emph{Dog}. Such hierarchical annotation makes it extremely difficult to precisely estimate the number of sound-sources in the clip. Hence, we propose to filter the original annotation by just keeping the third-level labels\footnote{Detailed filtering process can be found in the supplemental materials.}, e.g., \emph{Dog}.  In the original setting, all the videos are splitted into Evaluation/Balanced-Train/Unbalanced-Train set. For efficiency purpose, we only use the Balanced-Train set for training. These data are divided into different curriculums according to the number of contained sound-sources, e.g., the first curriculum $\mathcal{C}_1$ consists of videos with single sound-source. Finally, all the 19,443 valid video clips are divided into 9,239/7,098/2,685/421 $\mathcal{C}_1$/$\mathcal{C}_2$/$\mathcal{C}_3$/$\mathcal{C}_4$ curriculum clips.
For each curriculum set, the input audio is a 10s mono sound and the image is randomly selected from video.
Note that, all the semantic labels are not used during training.

\noindent{\textbf{MUSIC}}
The MIT MUSIC dataset contains 685 videos, with 536 musical solo and 149 duet videos. These videos contain 11 instrument categories.
They are also collected from YouTube, but cleaner than the ones in AudioSet. Hence, they are more proper for the sound separation task.
As the duet videos do not have ground-truths of sounds in mixtures, we only use the solo videos for training and testing.
Following \cite{gao2019co}, the first and second video of each instrument category is selected for validating and testing, respectively. And the rest ones in solo are for training.
Note that, some videos have been removed by the YouTube users, the final training data are about 467 videos. All the videos are randomly splitted into 10s clips.% which result in ...

\subsection{Network and Implementation Details}
Our audiovisual learning network is a two-stream network, where the off-the-shelf VGGish network~\cite{hershey2017cnn} is employed for sound and VGG16~\cite{simonyan2014very} is for vision. For each modality, the feature maps are the outputs of final convolution layer of the network. Detailed architecture description for both audiovisual learning model and separation model can be found in the supplemental materials.

For all the experiments, the input audio of 10s long is represented in log magnitude spectrogram of $512\times 432$, which is achieved by STFT (with the window size of 1022 and hop length of 256) and log-frequency projection. For the visual modality, we directly reshape the input image into $256\times 256 \times 3$. The network is trained with Adam optimizer, where the starting learning rate is $10^{-4}$ for the first curriculum, then gradually times $10^{-1}$ for the next one. For example, the learning rate for the third curriculum is $10^{-6}$.

%To effectively correlate these two modalities, we firstly reshape the output feature maps of each unimodal network into feature vectors, then embed them into a common space via a fully-connected layer for performing clustering and alignment.

% comparison method?

\subsection{Curriculum Learning Evaluation}
\subsubsection{Learning comparison}
\begin{figure}
\begin{minipage}[t]{0.5\linewidth}
\centering
\includegraphics[width=1.6in]{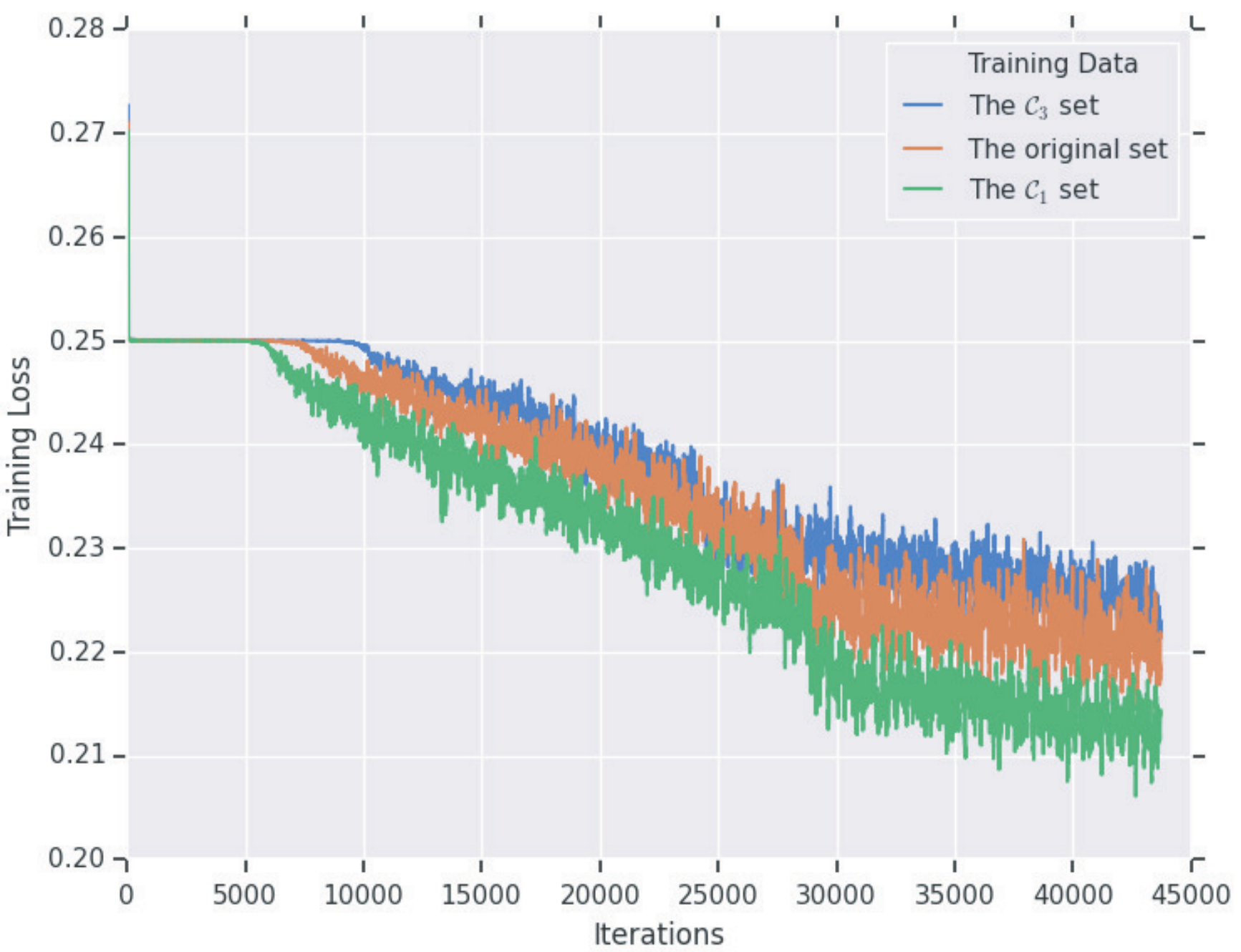}
\subcaption{Curriculum effects}
\label{fig:side:a}
\end{minipage}%
\begin{minipage}[t]{0.5\linewidth}
\centering
\includegraphics[width=1.6in]{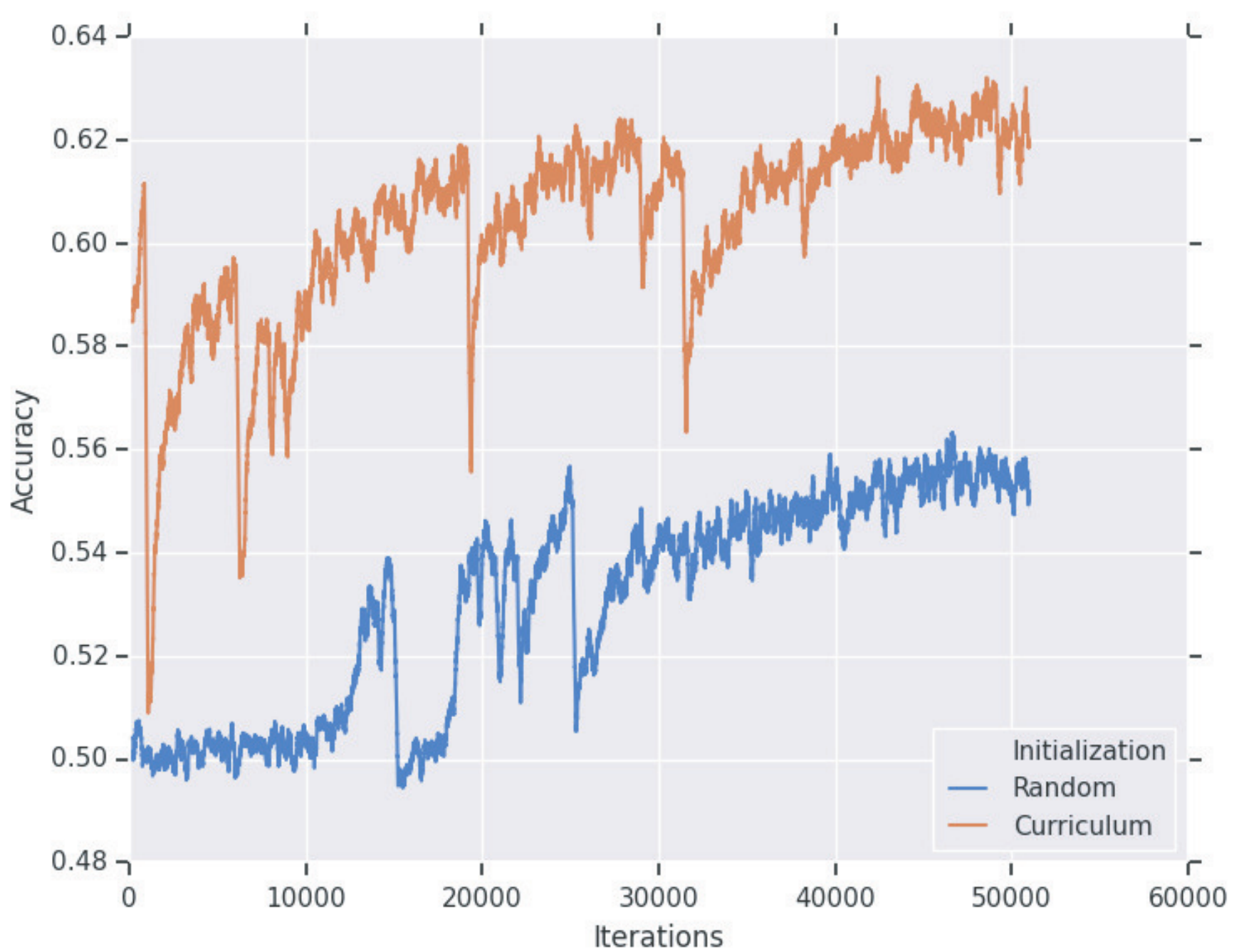}
\subcaption{Initialization effects}
\label{fig:side:b}
\end{minipage}
\caption{The effects of curriculum learning in terms of training loss and accuracy.}
\end{figure}

In this section, we aim to have an insight into how the curriculum strategy influences the audiovisual learning performance.
Concretely, to evaluate the effects of different audiovisual complexities, the original set and the curriculum set of $\mathcal{C}_1$ and $\mathcal{C}_3$ are selected for training the audiovisual network, respectively. As shown in Fig.~\ref{fig:side:a}, it is obvious that the network trained with the simple curriculum of $\mathcal{C}_1$ enjoys the fastest convergence and lowest training loss, while the one trained with $\mathcal{C}_3$ suffers from the worst performance. Such phenomena indicate that the model performance is significantly affected by the audiovisual complexity, and the simple scene can provide better learning performance.

Further, we want to know what the model can benefit from pre-curriculum, i.e., the effects of curriculum initialization. In Fig.~\ref{fig:side:b}, we show the training accuracy of audiovisual model on the $\mathcal{C}_3$ set, which are initialized from random and the model trained with the $\mathcal{C}_2$ set, respectively. Surprisingly,
the model initialized from $\mathcal{C}_2$  enjoys the great advantages compared with the random one. Curriculum learning indeed helps to accelerate and improve the audiovisual learning performance.

\subsubsection{Acoustic Scene Classification }
The unimodal representation learned by audiovisual model is also influenced by the curriculum strategy.
To assess such influence, we propose to perform acoustic scene classification by viewing the trained audiovisual model as a feature extractor.
The ESC-50 dataset is chosen for evaluation and we follow the same pre-processing and train/test split as \cite{hu2019deep}. Table~\ref{esc} show the comparison results, where the sound-source-level alignment is Eq.~\ref{eq5} and the scene-level alignment is directly comparing the audio and visual representation without clustering, similar to \cite{arandjelovic2017look}.  In Table~\ref{esc}, we can summarize these results into three points.
First, learning with simple curriculum can provide more proper initialization.
Second, sound-source matching better utilizes the audiovisual concurrency than scene-level matching, especially in the complex scene.
Third, direct video-level matching in the complex scene may deteriorate the pre-trained network. This is probably because the chaotic audiovisual correlation could confuse the scene matching objective, but it was ignored before.

\begin{table}
\begin{center}
\begin{tabular}{c|c}
\hline
Training Strategy &  Accuracy$\uparrow$ \\
\hline\hline
$\mathcal{C}_1$ &  51.25\\
$\mathcal{C}_2$+ Sound-source-level alignment & 56.75 \\
$\mathcal{C}_2$+ Scene-level alignment &  47.25\\
\hline
\end{tabular}
\end{center}
\vspace{-0.5cm}
\caption{Acoustic scene classification result on ESC-50.}\label{esc}
\end{table}

\subsubsection{Poisson Regression}
In this section, we evaluate the performance of audiovisual complexity estimation.
Table~\ref{poisson} shows the Poisson regression results when training on AudioSet-Balanced-Train and testing on AudioSet-Evaluation, where the number of sound source ranges from 1 to 5.
Compared to the chance results, the audio Poisson regression network has a great superiority in both accuracy and \emph{Mean Average Error} (MAE).
Moreover, the results can be further improved by adopting the network pre-trained with audiovisual learning objective, i.e., Eq.~\ref{eq5}.
Intuitively, if the network has been trained with higher level curriculum, e.g., $\mathcal{C}_2$, it will better estimate the number of sound-sources in the audio modality.
Such evidences show that complex audiovisual data can be effectively modeled by self-supervised learning method, especially when adopting the curriculum learning strategy.

\begin{table}
\begin{center}
\begin{tabular}{c|c|c}
\hline
Approach &  Accuracy$\uparrow$ & MAE$\downarrow$\\
\hline\hline
Random & 20.2 &1.624\\
w/o Pre-train  & 45.2  &0.742\\
CAVL-$\mathcal{C}_0$ &  47.0 & 0.701\\
CAVL-$\mathcal{C}_1$ &  49.6 & 0.616\\
CAVL-$\mathcal{C}_2$ &  50.9 & 0.614 \\
\hline
\end{tabular}
\end{center}
\vspace{-0.5cm}
\caption{Poisson regression results. CAVL-$\mathcal{C}_0$ is a preliminary curriculum, in which the model is trained with scene-level alignment on the $\mathcal{C}_1$ dataset.}\label{poisson}
\end{table}

% 10.29
\subsection{Audiovisual Sound Localization}
In this task, we aim to visualize the object location where the sound is produced. The AudioSet-Balanced-Train dataset is adopted for training, and the human-annotated SoundNet-Flickr~\cite{aytar2016soundnet,senocak2018learning} dataset is used for testing. As the training and testing datasets come from different sources, it is more challenging to perform exact localization.

\begin{figure}[t]
\begin{center}
   \includegraphics[width=0.85\linewidth]{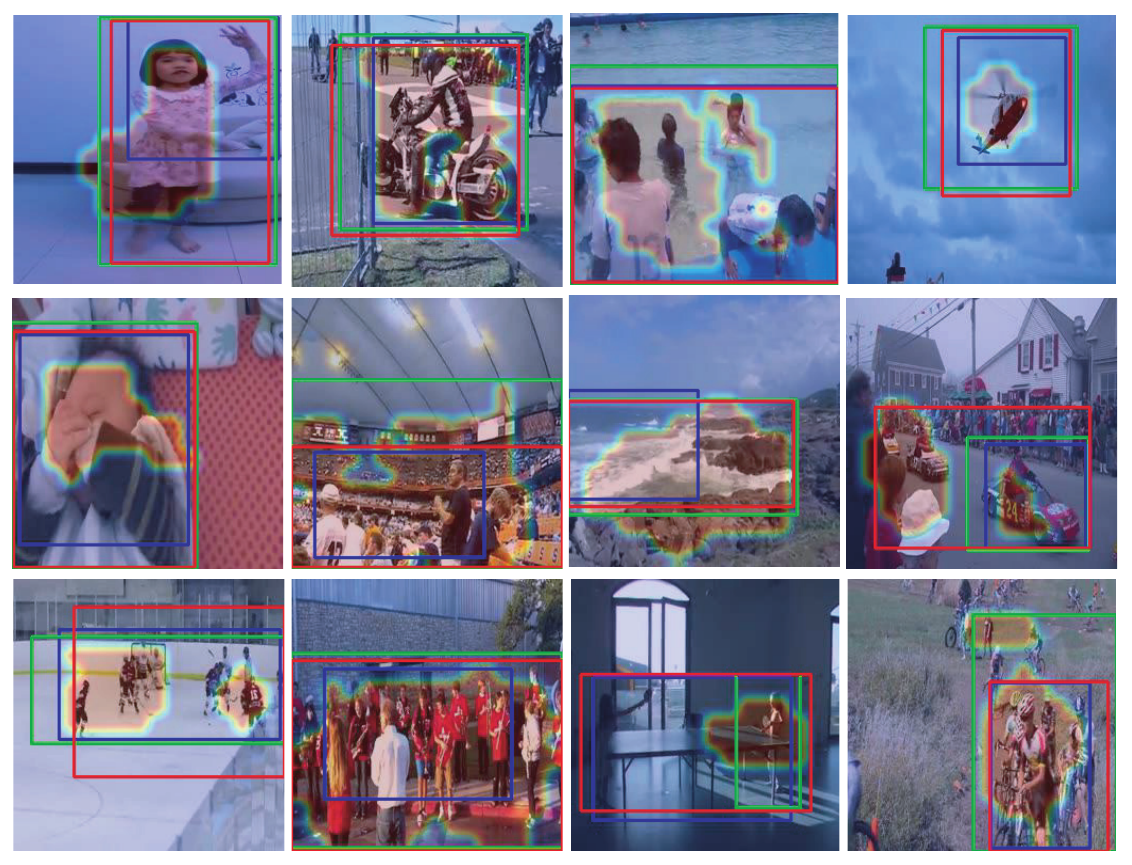}
\end{center}
\vspace{-0.5cm}
   \caption{Qualitative localization results on SoundNet-Flickr dataset~\cite{senocak2018learning}. The bounding boxes in different colors are the annotated source locations by different subjects. The assignments of our aligned visual centers are shown as heatmaps.}
\label{local}
\end{figure}
Fig.~\ref{local} shows some qualitative examples w.r.t. sound-source locations. Compared with the human-annotations, our model can predict proper visual localization for the corresponding sound. And we can find that the annotated bounding boxes are sometimes too rough to provide exact source locations, while our model can address this challenge due to the clustering advantage of spatial assignments.
%Besides, different from the
%different from \emph{Class Activation Map} (CAM) \ref{zhou2016learning}

We further perform quantitative evaluation. Following~\cite{senocak2018learning}, the same 250 audiovisual pairs are selected from the annotated SoundNet-Flickr dataset for evaluation, and consensus Intersection over Union (cIoU) and AUC area~\cite{senocak2018learning} are used as evaluation metrics. To evaluate the effectiveness of curriculum learning, our models trained in different curriculum levels are also considered, as shown in Fig.~\ref{location}.
First, our models outperform all the other methods by a large margin. It demonstrates that our model can better capture and align different sound-sources, even faced with multi-source scenes.
Second, besides the aligned visual center, we also evaluate the unaligned visual center. As expected, they suffer from a large decline in both metrics, which indicates that our model can exactly distinguish sound-maker from background and align it with the produced sound.
Third, our model trained with curriculum $\mathcal{C}_2$ is worse than the one with $\mathcal{C}_1$. This is because the test videos are all single-source, the multi-source videos in $\mathcal{C}_2$ may mix up the alignment knowledge learned in $\mathcal{C}_1$.

\begin{table}
\begin{center}
\begin{tabular}{c|c|c}
\hline
Methods &  cIoU@0.5$\uparrow$ & AUC$\uparrow$\\
\hline\hline
Random & 12.0 &32.3\\
Attention\cite{senocak2018learning} &  43.6 & 44.9\\
DMC\cite{hu2019deep} &  41.6& 45.2\\
CAVL-$\mathcal{C}_1$ &  50.0 & 49.2\\
CAVL-$\mathcal{C}_1$(unrelated) &  19.2 & 36.8\\
CAVL-$\mathcal{C}_2$ &  46.0 & 45.7 \\
\hline
\end{tabular}
\end{center}
\vspace{-0.5cm}
\caption{Quantitative localization results on SoundNet-Flickr dataset~\cite{senocak2018learning}. AUC is the area under the cIoU curve.}\label{location}
\end{table}

% 10.30
\subsection{Sound Separation}
% use the audiovisual sound separation task to evaluate the effectiveness of localized visual representation of sound-maker
We evaluate the audiovisual sound separation performance on the MIT-MUSIC dataset and more results on AudioSet are in the supplemental materials.
In this task, effective separation depends on the quality of visual representation of sound-maker.
To address this challenge, most of existing methods resort to the ImageNet pre-trained or fine-tuned sound-maker detector~\cite{gao2018learning,gao2019co,zhao2019sound,zhao2018sound}.
In contrast, we use the sound localization technique to automatically extract the visual representation of sound-maker, which is implemented with audiovisual alignment objective (i.e., Eq.~\ref{eq5}) without any human-supervision.
Fig.~\ref{music} shows some examples of solo and duet scene.
It is obvious that our model can localize most instruments, especially the different instruments in the duet scene. Then, we can use the corresponding visual centers or masked visual features as the representation of sound-maker for sound separation, as introduced in Sec.3.3.2.
Note that, as the extraction of visual representation does not use any extra human-annotation, it is more general and flexible than existing methods.
\begin{figure}[t]
\begin{center}
   \includegraphics[width=0.9\linewidth]{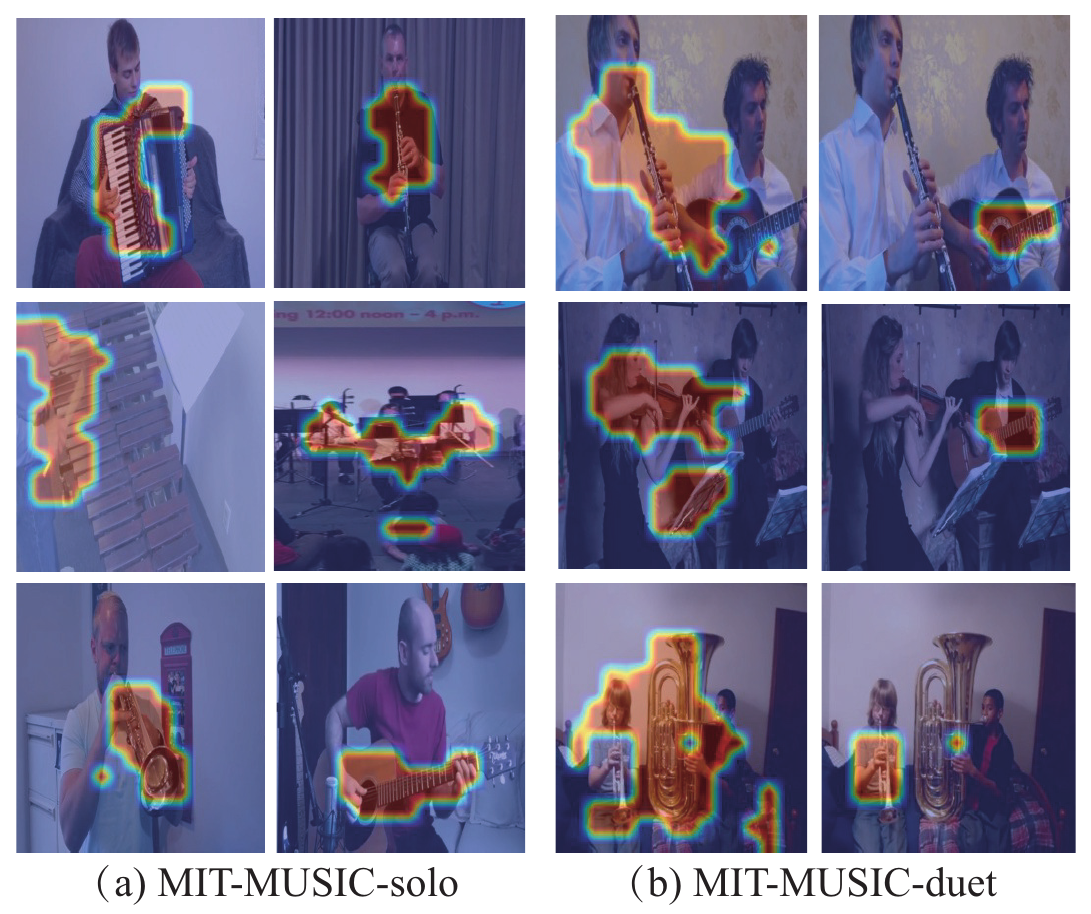}
\end{center}
\vspace{-0.5cm}
   \caption{Qualitative localization results on MIT-MUSIC dataset~\cite{zhao2018sound}.}
\label{music}
\end{figure}

In order to evaluate the separation results accurately, we use the synthetic mixture audios for evaluation, similar to~\cite{zhao2019sound,gao2019co}. The standard metrics of Signal-to-Distortion Ration (SDR), Signal-to-Interference Ratio (SIR), and Signal-to-Artifact Ratio (SAR) are used for evaluation.
Our model is compared with the audio-only separation method of NMF-MFCC~\cite{spiertz2009source} and the audiovisual separation models of AV-Mix-Sep~\cite{gao2019co}, Sound-of-Pixels~\cite{zhao2018sound} and Co-Separation~\cite{gao2019co}.
Table~\ref{solo} shows the separation results, where sound localization model is trained with solo videos.
Although the previous methods use additional visual knowledge for guiding the sound separation, e.g., ImageNet-pretrained visual model in AV-Mix-Sep~\cite{zhao2018sound} and finetuned instrument detector in Co-Separation~\cite{gao2019co}, our model still shows comparable results in both SDR and SIR\footnote{SAR measures the artifacts in the separated results instead of separation accuracy~\cite{gao2019co}.}.
Note that, our results are achieved with fewer training samples compared with others,
which demonstrates that our sound localization technique can provide effective visual representation of specific sound-maker.
Moreover, our separation model based on masked visual features performs much better than the one with visual center. This is probably because the masked visual features could provide more detailed representation of sound-maker than the aggregated one.
To validate the effectiveness of curriculum learning, the sound localization model is further trained with the duet videos. Table~\ref{duet} shows the ablation results. The performance gain in SDR and SAR indicates that our audiovisual alignment model can utilize complex scenes to improve the ability of cross-modal perception further.

% without explicitly utilizing the visual representation

\begin{table}
\begin{center}
\begin{tabular}{c|c|c|c}
\hline
Methods &  SDR$\uparrow$ & SIR$\uparrow$ & SAR$\uparrow$\\
\hline\hline
NMF-MFCC\cite{spiertz2009source} &  0.92 & 5.68 & 6.84\\
AV-Mix-Sep\cite{gao2019co} & 3.16 & 6.74 & 8.89\\
Sound-of-Pixels\cite{zhao2018sound} & 7.30 & 11.90 & 11.90\\
Co-Separation\cite{gao2019co} &  7.38 & 13.7 & 10.80 \\
Our-Model-Center &  5.79 &  9.15& 12.29\\
Our-Model-Mask &  6.59 & 10.10 &  12.56\\
\hline
\end{tabular}
\end{center}
\caption{Sound separation results on MIT-MUSIC-solo test set. All the methods are trained only with solo videos.}\label{solo}
\end{table}

\begin{table}
\begin{center}
\begin{tabular}{c|c|c|c}
\hline
Methods &  SDR$\uparrow$ & SIR$\uparrow$ & SAR$\uparrow$\\
\hline\hline
Our-Model-solo &  6.59 & 10.10 &  12.56\\
Our-Model-both &  6.78 & 10.62 &  12.19\\
\hline
\end{tabular}
\end{center}
\caption{Sound separation results on MIT-MUSIC-solo test dataset, where Our-Model-solo is trained only with solo videos, while Our-Model-both is trained with both solo and duet videos.}\label{duet}
\end{table}

% 10.31
\section{Conclusion}
In this paper, we developed an audiovisual learning model that discovers, then aligns the sounds and sound-makers in arbitrary audiovisual scenes. A curriculum learning strategy is proposed to effectively train the model w.r.t. the number of sound-source. Further, we deployed the well-trained audiovisual model into practical perception tasks. We achieved noticeable audiovisual localization performance, and the localized object representation made a considerable boost to sound separation.

{\small
\bibliographystyle{ieee_fullname}
\bibliography{cavl}
}

\clearpage

\section{Supplemental Materials }
\subsection{Curriculum Settings}
% filtering
The audio event dataset of AudioSet is annotated via a hierarchical ontology. For example, the sound of barking (the fourth-level label) is simultaneously annotated as \emph{Animal} (the first-level label), \emph{Pets}(the second-level label), and \emph{Dog}(the third-level label), i.e.,  Animal $>$ Pets $>$ Dog $>$ Bark.
Such hierarchical annotation makes it extremely difficult to precisely estimate the number of sound-sources in the clip. Considering that the third-level labels generally relate to the common objects in our surroundings, we propose to filter the original annotation by just keeping the third-level ones.
This filtering process consists of two steps, i.e., removing the father annotations and reducing the children annotations.
Concretely, for each third-level label of the video clip, we remove all its father annotations (the first and second label) if they appear in the label list of the same video clip. Meanwhile, for each fourth-level label, we reduce them into their father annotation of the corresponding third-level label then remove them. For example, we can find that the third-level label for \emph{Bark} is \emph{Dog} via the hierarchical ontology, then we use \emph{Dog} to replace \emph{Bark}.
After finishing these two steps, the annotation list for each video clip should only contain the labels in the third-level. Finally, we can use the number of labels as the indicator of the number of sound-sources.

\subsection{Audiovisual Learning Network}
The audiovisual learning network is a two-stream network, which consists of a audio pathway and a visual pathway.
We use the off-the-shelf VGGish network for the audio pathway, where we discard its last three fully-connected layers and last max-pooling layer. The output of the audio pathway is the feature maps of the final convolution layer of VGGish, with the size of $64 \times 54 \times 512$.
The similar procedure is performed with the visual pathway, but the model is replaced with VGG16, and the size of output visual feature maps is $16 \times 16 \times 512$.

For the output features of both modalities, we use the \emph{Reshape} operator to transform them into a set of vectors, i.e., reshaping from $64 \times 54 \times 512$ to $3456 \times 512$ for audio and from $16 \times 16 \times 512$ to $256 \times 512$ for visual modality. We use a fully-connected layer of $512-512$ to encode these reshaped features in the channel space. Then, the modality-specific clustering module is performed, with which we expect to discover concrete audio and visual contents. And the contrastive loss is used to train the whole network.

\subsection{Poisson Regression Network}
The Poisson regression network is developed based on the audio network of VGGish. We remove the last three fully-connected layers of VGGish but use \emph{GlobalMaxPooling} over the output feature maps to obtain a $512D$ feature vector. Then, two fully-connected layers of $512-512-1$ are employed to project the feature vector into a predicted value for the Poisson average value of $\lambda$. Finally, we can train the regression network w.r.t. the Poisson regression loss.
The SGD optimizer is used, whose momentum is set to $0.9$ and the learning rate is initialized at $1e^{-2}$ and decays by $0.01/(1+epoch*0.5)$.

\subsection{Sound Separation Network}
Our audiovisual sound separation network consists of two parts, one is for visual representation extraction of sound-maker and the other is for sound separation. Concretely, the visual branch is basically the audiovisual learning network. It takes image and corresponding sound as inputs, and extracts the visual representation of specific sound-maker in the scene. To automate this process, we use the audiovisual scenes with single sound-source. This is because we can directly localize the sound-maker by comparing different visual representations with the unique sound and without manual distinguishment. Then, we can use the corresponding clustering center as the representation of the localized sound-maker, which is a $512D$ vector.

The sound separation branch is a variant of U-Net\cite{ronneberger2015u}, which consists of an encoder and a decoder.
The encoder contains 6 convolution layers with $16-32-64-128-256-512$ channels, while the decoder contains 6 up-convolution layers with $512-256-128-64-32-16$ channels. The convolution layers in the encoder use $4 \times 4$ filters, and followed by a BatchNormalization and a LeakyReLu (with a slope of $0.2$) layer. The up-convolution layers in the decoder also use $4 \times 4$ filters, but followed by a BatchNormalization and a ReLu layer. The last up-convolution layer is followed by a sigmoid function to match value of the spectrogram mask.
Similar to the original U-net, we apply skip-connection between symmetric encoder and decoder layers.

To integrate the above two branches, the $512D$ visual representation is first passed to a fully-connected layer of $512-512$, then a BatchNormalization and a LeakyReLU  (with a slope of $0.2$) layer. The resulted $512D$ visual vector is replicated to match the size of encoded audio feature maps. Then, the audio and visual feature maps are concatenated together, and fed to the decoder of the sound separation branch.

\subsection{More results}
\subsubsection{Sound Separation}
\noindent
\textbf{AudioSet-Instrument} Following \cite{gao2018learning}, all the clips in AudioSet\cite{gemmeke2017audio} are filtered to construct a subset of 15 musical instruments. The filtered clips from the Unbalanced-Train set constitute the training dataset, the ones from Balanced-Train set are splitted for validation and testing. As some video clips in AudioSet have been removed by the uploader, the whole instrument dataset is smaller than the ones in \cite{gao2019co}, which is about 99,882/456/456 train/val/test clips.
As 93,679 video clips in the training dataset have single sound-source, we directly use them for training the audiovisual learning model.
Then, we use the well-trained model for extracting the visual representation of localized sound-source, with which we train the separation model, as shown in Fig.~\ref{music}.

Table.~\ref{separation} shows the sound separation results. We can find that our model shows superior performance over most compared methods, even some of them adopt additional visual knowledge. For example, AV-MIML~\cite{gao2018learning} and Sound-of-Pixels~\cite{zhao2018sound} use ImageNet pre-trained model as the visual extractor. Besides, we also show the results of CAVL-$\mathcal{C}_1$. Such results are obtained by directly viewing the audio assignment in the audiovisual learning model as the spectrogram mask. However, as the audio assignment is achieved over the embedded feature maps, it is too coarse to perform fine-grained spectrogram prediction. Hence, such method does not work for sound separation.

\begin{table}[h]
\begin{center}
\begin{tabular}{c|c|c|c}
\hline
Methods &  SDR & SIR & SAR\\
\hline\hline
Sound-of-Pixels\cite{zhao2018sound} & 1.66 & 3.58 & 11.5\\
AV-MIML\cite{gao2018learning} & 1.83 & - & -\\
NMF-MFCC\cite{spiertz2009source} &  0.25 & 4.19 & 5.78\\
Co-Separation\cite{gao2019co} &  3.65 & 6.13 & 13.2 \\
CAVL-$\mathcal{C}_1$ &  -4.54 & 0.23 & 2.28\\
Our-Model-Center &  2.40 &  3.33& 18.24\\
Our-Model-Mask &  2.64 & 3.45 &  17.17\\
\hline
\end{tabular}
\end{center}
\caption{Sound separation results on AudioSet-Instrument test set.}\label{separation}
\end{table}

\begin{figure}[!tbp]
\begin{center}
   \includegraphics[width=0.9\linewidth]{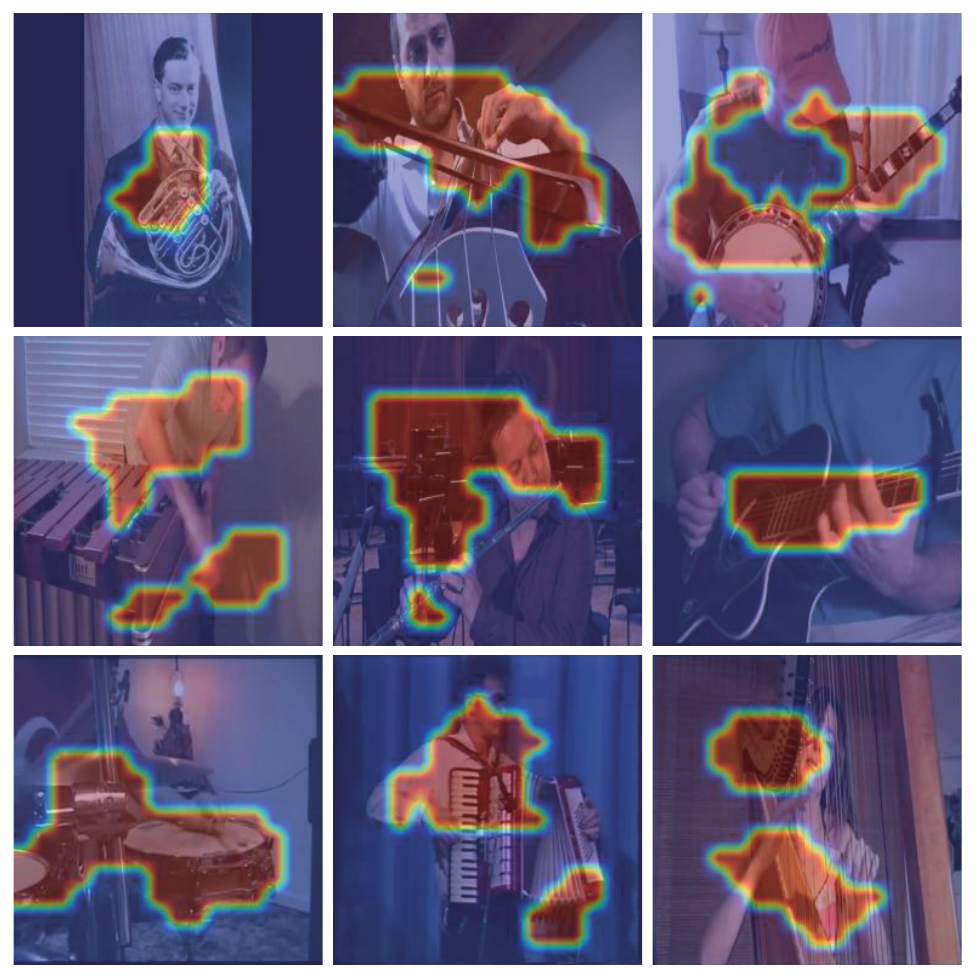}
\end{center}
   \caption{Qualitative localization results on AudioSet-Instrument test dataset.}
\label{music}
\end{figure}

\end{document}